\def\paperID{8303}
\def\confName{CVPR}
\def\confYear{2024}

\def\paperTitle{LightOctree: Lightweight 3D Spatially-Coherent Indoor Lighting Estimation}

\def\authorBlock{
    Xuecan Wang\textsuperscript{1}\thanks{Equal contribution} \qquad
    Shibang Xiao\textsuperscript{1}\footnotemark[1] \qquad
    Xiaohui Liang\textsuperscript{1,2}\thanks{Corresponding author} \\
    \textsuperscript{1} Beihang University, China \\
    \textsuperscript{2} Zhongguancun Laboratory, China \\
    {\tt\small \{xuecan\_wang, xiaoshibang, liang\_xiaohui\}@buaa.edu.cn}
}

\newif\ifreview 
\newif\ifarxiv 
\newif\ifcamera \newcommand{\cameraready}{\cameratrue}
\newif\ifrebuttal 

\cameraready

\pdfoutput=1
\documentclass[10pt,twocolumn,letterpaper]{article}
\ifreview \usepackage[review]{cvpr} \fi
\ifarxiv \usepackage[pagenumbers]{cvpr} \fi
\ifrebuttal \usepackage[rebuttal]{cvpr} \fi
\ifcamera \usepackage{cvpr} \fi

\usepackage{graphicx}	
\usepackage{amsmath}	
\usepackage{amssymb}	
\usepackage{booktabs}
\usepackage{times}
\usepackage{microtype}
\usepackage{epsfig}
\usepackage[table,xcdraw,dvipsnames]{xcolor}
\usepackage{caption}
\usepackage{float}
\usepackage{placeins}
\usepackage{color, colortbl}
\usepackage{stfloats}
\usepackage{enumitem}
\usepackage{tabularx}
\usepackage{xstring}
\usepackage{multirow}
\usepackage{xspace}
\usepackage{url}
\usepackage{subcaption}
\usepackage{xcolor}
\usepackage[hang,flushmargin]{footmisc}

\usepackage{cases}
\usepackage{algorithm}
\usepackage{algorithmic}

\ifcamera \usepackage[accsupp]{axessibility} \fi

% ...

\ifarxiv  \fi

\newcommand{\R}[1]{{%
    \textbf{%
        \ifstrequal{#1}{1}{\textcolor{red}{R#1}}{%
        \ifstrequal{#1}{2}{\textcolor{blue}{R#1}}{%
        \ifstrequal{#1}{3}{\textcolor{magenta}{R#1}}{%
        \ifstrequal{#1}{4}{\textcolor{teal}{R#1}}{%
                           \textcolor{cyan}{R#1}%
        }}}}%
    }%
}}

\usepackage{xr-hyper}

\makeatletter
\newcommand*{\addFileDependency}[1]{
  \typeout{(#1)}
  \@addtofilelist{#1}
  \IfFileExists{#1}{}{\typeout{No file #1.}}
}

\makeatother

\definecolor{cvprblue}{rgb}{0.21,0.49,0.74}
\usepackage[pagebackref,breaklinks,colorlinks,citecolor=cvprblue]{hyperref}
\usepackage[capitalize]{cleveref}
\crefname{section}{Sec.}{Secs.}
\crefname{table}{Table}{Tables}
\crefname{figure}{Fig.}{Figs.}

\definecolor{myblue}{RGB}{68, 114, 196}
\definecolor{mygreen}{RGB}{112, 173, 71}
\definecolor{myyellow}{RGB}{255, 192, 0}

\begin{document}
\title{\paperTitle}
\author{\authorBlock}
\maketitle

\begin{abstract}
    We present a lightweight solution for estimating spatially-coherent indoor lighting from a single RGB image. 
    Previous methods for estimating illumination using volumetric representations have overlooked the sparse distribution of light sources in space, necessitating substantial memory and computational resources for achieving high-quality results. 
    We introduce a unified, voxel octree-based illumination estimation framework to produce 3D spatially-coherent lighting. 
    Additionally, a differentiable voxel octree cone tracing rendering layer is proposed to eliminate regular volumetric representation throughout the entire process and ensure the retention of features across different frequency domains.
    This reduction significantly decreases spatial usage and required floating-point operations without substantially compromising precision.
    Experimental results demonstrate that our approach achieves high-quality coherent estimation with minimal cost compared to previous methods.
    
\end{abstract}

\section{Introduction}
\label{sec:intro}


In the realm of mixed reality and image editing, achieving visual coherence is a fundamental research concern when integrating virtual objects into real-world images. Lighting is particularly crucial for seamlessly blending virtual and real elements, ensuring consistent shading and shadowing between virtual objects and the surrounding environment. However, estimating the inverse process of global environmental lighting from narrow field-of-view (FOV) images is a challenging task due to the lack of sufficient information. 
\begin{figure}[h]
    \centering
    \includegraphics[width=\linewidth]{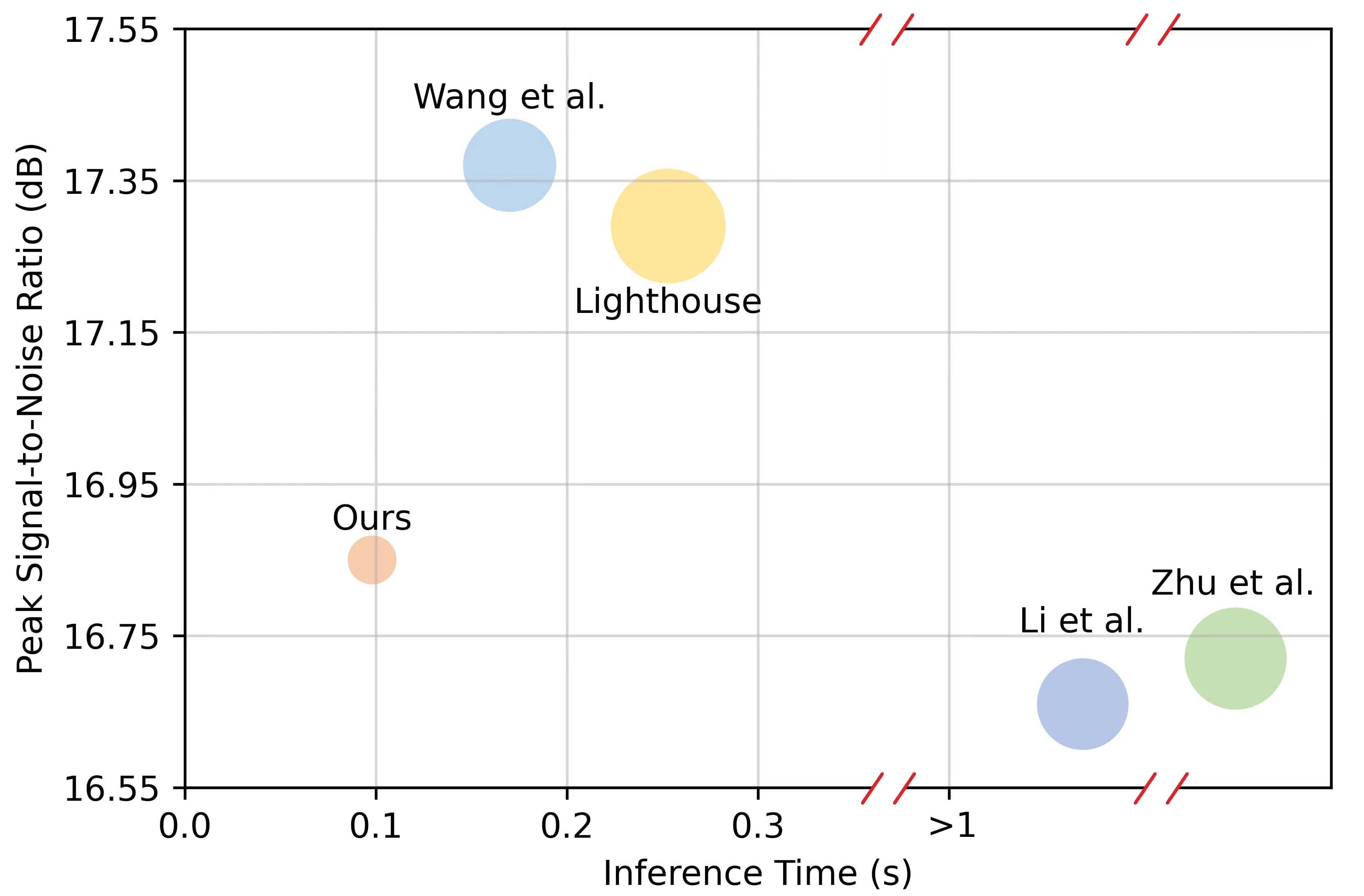}
    \caption{The comparison with state-of-the-art (SOTA) methods shows that our approach achieves relatively good accuracy while requiring lower storage and computational costs.}
    \label{fig:experiment_lightweight}
\end{figure}
Particularly in indoor environments, the proximity of lights to objects typically leads to significant variations in incident illumination across different positions within the scene. Moreover, when incorporating moving objects into the scene, ensuring lighting coherence during spatial changes is crucial. This necessitates the estimation of a spatially-coherent model of illumination that is 3D spatially-coherent \cite{PratulP.Srinivasan.2020}. Given these requirements and application scenarios, it is essential for the entire process of lighting estimation and virtual-real fusion rendering to be efficient and lightweight.

Some of existing work generate 2D illumination map as global lighting estimation. Such as methods using light probes \cite{Debevec_Graham_Busch_Bolas_2012, Unger_Gustavson_Ynnerman_2007, Löw_Ynnerman_Larsson_Unger_2009}, or deep neural networks like 2D CNNs \cite{Barron.2015, MarcAndreGardner.2017} to make predictions in image space.
However, these algorithms currently used to estimate global illumination either ignore spatial variation effects and predict a single illumination for the entire scene \cite{YeYu.2019, SoumyadipSengupta.2019, ChloeLeGendre.2019, Chalmers_Zhao_Medeiros_Rhee_2021, FangnengZhan.2021, Xu.2022, Wang.2022}, or estimate spatially-varying illumination by separately predicting the lighting at each 3D position in the scene \cite{MathieuGaron.2019, ZhengqinLi.2020, Zhao.2021, Zhu_Li_Matai_Porikli_Chandraker_2022, Liu_Wang_Li_Quan_Xu_2023}. 
Although these approaches may produce impressive results, they cannot guarantee smooth variations in predicted illumination with changes in position, especially when inserting moving virtual objects, which can easily break the illusion. 
Therefore, recent research has explored illumination estimation methods based on extending the lighting representation to three-dimensional space, typically using 3D volumetric $RGB\alpha$ models \cite{PratulP.Srinivasan.2020, ZianWang.2021, JingsenZhu2022LearningbasedIR}. 
These methods facilitate spatially-coherent lighting estimation, but they also introduce higher-dimensional lighting representations and more complex computations. Such as 3D CNNs, which are typically computationally and memory-intensive. 
At the same time, they overlook the sparse distribution of lighting in the environment, leading to significant waste of space and computing power, making them unsuitable for performance-constrained applications.

Upon observation, we note that assuming the incident radiation at a certain point comes solely from the surfaces in the scene, a common assumption in existing methods, leads to a sparse distribution of surfaces in three-dimensional space. Therefore, we can use a sparse data structure as a lighting representation instead of a coarse uniform grid. Drawing inspiration from NeRFs\cite{Keil_Yu_Tancik_Chen_Recht_Kanazawa_2022, Kerbl_Kopanas_Leimk_2023,Yu_Li_Tancik_Li_Ng_Kanazawa_2021} and octree-based global illumination researches\cite{Crassin_Neyret_Sainz_Green_Eisemann_Llaguno_2011, laine2010efficient}, our approach delves into a lighting representation based on sparse voxel octrees and proposed a lightweight, spatially-coherent global lighting estimation network that accounts for the distribution characteristics of the light field in the scene. We also introduce a differentiable rendering layer based on the voxel octree representation to streamline the framework and better align with the training of the octree-based network. Additionally, we explore a rendering method for virtual object insertion based on a hybrid scene representation using voxel octrees and point clouds. 
By restricting data storage and calculations to octants, our method incur a memory and computational cost of $O(n^2)$, where $n$ is the voxel resolution per dimension at the finest granularity level. In contrast, utilizing a 3D uniform voxel grid representation solution results in a memory and computational cost of $O(n^3)$. This approach enables low-cost, high-quality coherent augmentations.

In summary, our main contributions are as follows:
\begin{itemize}
\item An octree-based framework to achieve virtual-real lighting consistency, optimized for less memory and hardware requirements with minimal loss of accuracy. This framework efficiently generates 3D spatially-coherent lighting interactively and inserts virtual objects in real-time using only a single RGB image of indoor scenes.

\item A novel lighting representation base on voxel octree for indoor lighting estimation, which considered the sparsity of light field in 3D space, significantly reducing storage space and computational complexity.

\item A lightweight lighting estimation network, featuring a novel multi-scale rendering layer. This network enables end-to-end estimation high-quality incident radiance fields in the form of the voxel octree.

\end{itemize}

\section{Related Work}
\label{sec:related}

To achieve coherent augmentations, three primary issues must be addressed: geometric registration, photometric registration, and camera simulation \cite{Schmalstieg_Hollerer_2017}. This paper specifically focuses on photometric registration, which involves the interaction of light between the real world and the augmented effects. 
In this section, we will provide a brief review of pertinent prior research endeavors focused on estimating diverse lighting representations from images, leveraging the simplifying assumptions of the light field.

\vspace{-0.25cm}
\paragraph{Non Spatically-varying Lighting Estimation.} 
Early research assumed that the incident light field received at any position in the scene is uniform, simplifying the spatial distribution of illumination to a single spherical distribution. In this scenario, the scene’s lighting can be represented by a HDR panorama. This conclusion was validated by Debevec et al.\cite{Debevec_2008}, who demonstrated that virtual objects can be realistically inserted into real photographs using HDR environment maps obtained through multiple exposures of a chrome probe. Subsequent methods \cite{Barron.2015, MarcAndreGardner.2017, YeYu.2019, SoumyadipSengupta.2019, ChloeLeGendre.2019, Chalmers_Zhao_Medeiros_Rhee_2021, FangnengZhan.2021, Xu.2022, Wang.2022} have illustrated that deep learning techniques can estimate HDR environment maps from a single LDR photograph. These methods offer the advantage of significantly reducing the dimensionality of the output space of the lighting estimation algorithm, thereby enabling the design of complex network models to accomplish more detailed image generation tasks. However, a single environment map is insufficient for compositing multiple, large, or moving virtual objects into the captured scene\cite{PratulP.Srinivasan.2020}, particularly in indoor environments where light sources and other scene content may be in close proximity to the insertion positions of the objects.

\vspace{-0.25cm}
\paragraph{Spatically-varying Lighting Estimation.} 
To overcome the limitations of single environment map methods, recent research has delved into spatially-varying lighting estimation algorithms for indoor scenes. Numerous studies have predicted spatially-varying lighting in images by estimating per-pixel spherical lobes \cite{Gardner.20191020, RuiZhu.2022, ZhengqinLi.2020, ZhengqinLi.2021, ZhengqinLi.2022} (spherical harmonics/Gaussians) or individual environment maps for each pixel \cite{Liu_Wang_Li_Quan_Xu_2023, MathieuGaron.2019} in the input image. However, these methods overlook the depth factor and can only estimate the lighting for the surface corresponding to any given pixel in the image. Consequently, the estimated lighting lacks continuity in three-dimensional space, as highlighted by Srinivasan et al.\cite{PratulP.Srinivasan.2020} and Karsch et al.\cite{Karsch.2014}. In the most recent work\cite{PratulP.Srinivasan.2020, ZianWang.2021, JingsenZhu2022LearningbasedIR}, higher-order voxelized lighting representations were utilized for estimation, and a 3D neural network was employed to achieve spatially-coherent lighting estimation. Nevertheless, the use of 3D networks and 3D lighting representations significantly increases the storage and computational complexity of the algorithm, hindering its application in AR. Additionally, in general indoor scenes where special objects like participating media are not considered, the incident radiance at a point can be simplified to only come from the surfaces in the scene. Therefore, using the 3D voxel grid structure of existing methods to represent lighting would result in unnecessary storage and computational losses. Our approach explores the use of voxel octree as a compressed representation for lighting/scene and designs a lightweight and spatially-coherent global lighting estimation network. By restricting network and rendering to an octree, and combining them with a hierarchical feature fusion rendering layer design, we achieve low-cost full-frequency global lighting estimation from a single RGB image.

\begin{figure*}[htb]
    \centering 
    \includegraphics[width=2.1\columnwidth]{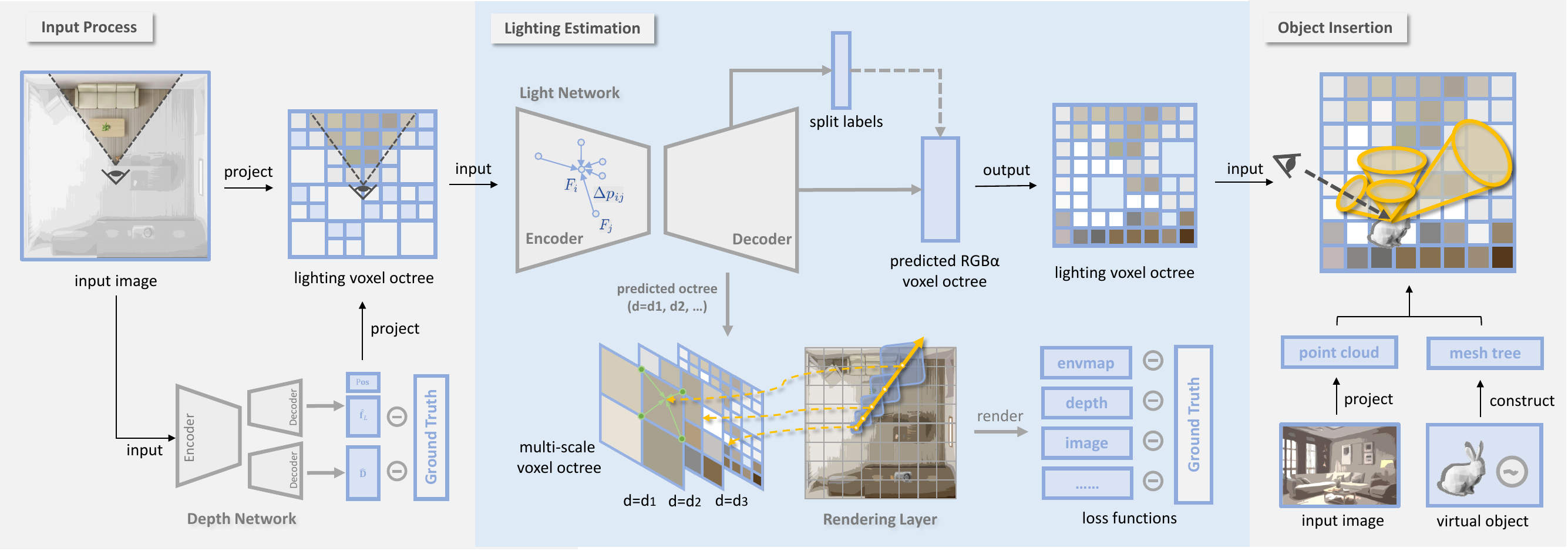}
    \caption{\textbf{Overall structure of our framework.} The process starts with the Input Process stage, where the direct prediction module estimates depth and extracts global illumination features from the input image. The depth values and features are combined to create a point cloud, which is used to build a 3D voxel octree scene representation. Next, the Lighting Estimation stage uses a U-Net structure to predict the lighting from the constructed octree. The Object Insertion stage then combines the original RGB information, predicted depth values, predicted lighting voxel octree, and user-specified mesh data to render an image with consistent lighting for the inserted virtuals object.}
    \label{fig:overview_pipeline}
   \end{figure*}
   
   \section{Overview}
   Given a single LDR RGB image in narrow FOV, we aim to estimate a 3D voxel octree lighting representation of the scene with minimal storage and computational cost, and integrate the estimation results with information about virtual objects. 
   Our pipeline can be divided into three stages, as shown in Fig.\ref{fig:overview_pipeline}, including two neural network and one object rendering module. In the follow sections, we will illustrate how our voxel octree is involved throughout the entire workflow to reduce the cost of storage and computation.
   
   In Sec.\ref{sec:LightweightLightingEstimation}, we will describe our lightweight illumination estimation method and corresponding training strategy, covering the input process and the lighting estimation stages. 
   In Sec. \ref{sec:experiment}, we will showcase several experiments conducted to validate our designs. Finally,  Sec.\ref{sec:conclusion} will provide a brief summary and outlook of our approach.

\section{Lightweight Lighting Estimation} \label{sec:LightweightLightingEstimation}

\subsection{Depth Network}
The Depth Network is intended to provide initial predictions of depth and global features including lighting information when given a single image as input. 
Adopting a similar approach to \cite{JingsenZhu2022LearningbasedIR}, we employ DenseNet121 \cite{Maaten_Weinberger_2017} as the backbone for the 2D CNN encoder. The depth decoder utilizes skip connections to generate the final predicted depth $\hat{D}\in \mathbb{R}^{3\times H \times W}$. Moreover, drawing inspiration from the work in \cite{ZianWang.2021}, we incorporate an additional module subsequent to the initial encoder to extract global lighting information, which is ultimately output as a feature vector $\hat{\mathbf{f}_L} \in \mathbb{R}^{C}$. These features are then amalgamated with the input of the Light Network to facilitate the prediction process.

\subsection{Light Network}
The Light Network is to predict lighting octree from RGB-D values and global lighting features. To achieve lightweight lighting estimation, a compact lighting representation supporting fast, parallel network computation and rendering is essential. Thus, we devised a lighting voxel octree, a 3D volumetric lighting representation organized using an octree. The octree efficiently represents the surface radiance exiting from the entire scene, encompassing both visible surfaces and those outside the FoV. However, introducing the octree has presented significant challenges for network design, particularly for the ill-posed regression task of lighting estimation based on a single image, which previous neural network models using octree representation could not address. To tackle this issue, we integrated the hierarchical structure of the octree with a rendering layer based on differentiable cone tracing, facilitating the fusion of multi-scale lighting features. This enables us to maximize the network’s receptive field while ensuring its preference for inductive lighting estimation at minimal cost. Subsequently, we developed a lightweight lighting estimation network based on octree graph network operators. The following section provides a comprehensive description of the model design and training specifics for this segment.

\begin{figure*}
    \begin{minipage}{.7\textwidth}
	\centering	
        \includegraphics[width=\textwidth]{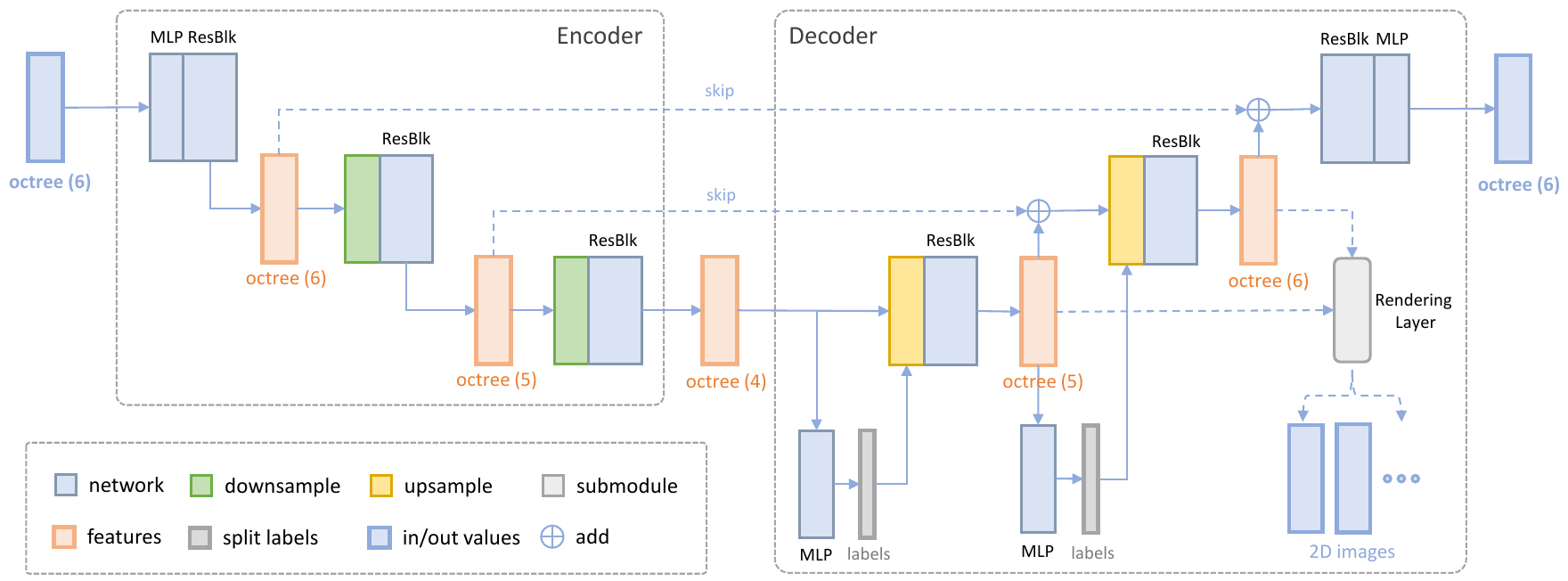}
    \end{minipage}
    \hspace{0.025pt}
    \begin{minipage}{.285\textwidth} 
        \vspace{0.05cm}
        \captionof{figure}{\textbf{Light Network.} A lightweight U-Net architecture constructed with graph O-CNN operators. These modules work in tandem to forecast octree subdivisions, field values, and establish correspondence with ground truth through the Rendering Layer.To provide a clearer illustration of the structure, here presents a network with a full depth of 4 and depth of 6, while the actual depth is 7 (equivalent to voxel resolution of $128^3$). }
        
        \label{fig:network_design}
    \end{minipage}
\end{figure*}

\subsubsection{Lighting Presentation}

\paragraph{Lighting Voxel Octree.} We assumes that the lighting in the scene emanates solely from the object surfaces, including light sources and other objects, by simplifying the image formation process into a surface rendering model. Based on this characteristic, a 3D voxel octree structure is adopted to tightly store the scene and lighting. The voxel octree, developed as a sparse lighting representation, has eight child nodes for each non-empty non-leaf node. Each non-empty node stores its corresponding shuffle key, split label, and features. For illumination, the feature information stored in each node is the RGB$\alpha$ value, representing the radiance emitted from that position. Meanwhile, each empty node only stores the shuffle key and label. This structure is compatible with the data structure used in \cite{Wang_Liu_Tong_2022} and is utilized to build and design the lighting prediction network using basic modules such as octree-based neural modules proposed by \cite{PengShuaiWang2017OCNNOC, Wang_Liu_Tong_2020, Wang_Liu_Tong_2022}.

\vspace{-0.25cm}
\paragraph{Octree Construction.} When provided with depth values and corresponding features (\textit{i.e.} RGB values)
, and the camera intrinsic matrix, a 3D point cloud of thus region can be obtained through projection transformation, denoted as $\mathcal{P} = \{(\mathbf{p}_1, \mathbf{f}_1), (\mathbf{p}_2, \mathbf{f}_2), \ldots, (\mathbf{p}_N, \mathbf{f}_N)\}$. Here, $\mathbf{p}_i$ represents the position information of point $i$, which is the 3D coordinate $(x_i, y_i, z_i)$ in the camera coordinate system, and $\mathbf{f}_i$ represents the feature information of the point, \textit{i.e.} the color information $(r_i, g_i, b_i)$ corresponding to the RGB color value of the point. To construct the octree of the input 3D point cloud data, the point cloud is initially proportionally scaled to an axis-aligned unit bounding box. Subsequently, the 3D voxel grid data undergoes recursive subdivision in breadth-first order until the predefined octree depth $d$ is reached. At each step, all non-empty octree nodes which is occupied at the current depth $l$ are traversed, and they are subdivided into eight child octree nodes at the next depth $l+1$.

\subsubsection{Network Architecture}

Given an RGB image and pixel-wise depth as input, we start by constructing a voxel octree representing the visible area. Subsequently, we incorporate the global illumination feature information $\mathbf{f}_L$ and positional data, which is stochastically generated based on the input depth distribution. This leads to an update in the octree structure and features. 
Following this, utilizing the voxel octree structure and a series of neural modules based on the octree data structure \cite{PengShuaiWang2017OCNNOC, Wang_Liu_Tong_2020, Wang_Liu_Tong_2022}, we design a network model for light estimation task, as illustrated in Fig.\ref{fig:network_design}. This model employs the U-Net \cite{CicekALBR16} structure:
\begin{equation}
    \hat{L} = h_{LN}(I,D,\mathbf{f}_L;\Theta_{LN})
    \label{eq:light_network}
\end{equation}
$\mathrm{ResBlock}(n,c)$ in Fig.\ref{fig:network_design} represents a stack of $n$ ResNet blocks \cite{He_Zhang_Ren_Sun_2016}, each comprising two graph convolutions with a channel number of $c$. Moreover, $\mathrm{DownSample}(c)$ and $\mathrm{UpSample}(c)$ refer to graph downsampling and upsampling operations based on shared fully connected layers, with both input and output channels set to $c$. At the end of each stage in the decoder, the extracted features are input into the $\mathrm{PredictionModule}$ to predict the subdivision of octree nodes. This module consists of an MLP with two fully connected layers. Additionally, the extracted features are passed to the $\mathrm{RegressionModule}$ to predict the field values corresponding to the octree nodes in that layer. Subsequently, features stored in different layers of the voxel octree, along with camera poses, are sent to the $\mathrm{RenderingLayer}$ to render novel view images and calculate losses with ground truths.

\subsubsection{Rendering Layer}
Leveraging multi-level voxel octree for illumination representation, we propose a differentiable cone tracing rendering layer. This module efficiently conducts rendering and provides supervision for the network. Building upon the octree-based illumination representation, it rapidly renders panoramic images while performing mip-map sampling based on the distance between octants and camera. It exhibits high sensitivity to nearby light octants, resulting in high-frequency rendering effects. Simultaneously, it aggregates distant light sources, reducing sampling frequency to enhance speed without compromising the final rendering quality. Furthermore, we derive the differentiable form of this method, enabling it to serve as a gradient-propagating rendering layer within neural networks.

The conventional volume rendering process calculates the light radiation $L(\mathbf{x}, \omega)$ from viewpoint $\mathbf{x}$ in direction $\omega$, factoring in the sampling distance $\delta_k$ between sampling points $n-1$ and $n$. This process can often lead to redundant computations and complicate the efficient integration of feature information across different scales throughout the rendering process. However, our method not only streamlines the rendering process but also skillfully facilitates feature fusion across the outputs of different network layers.

In particular, when intending to render a panoramic image of the target camera pose from the predicted lighting voxel octree, we first generate cones that cover the entire 360-degree sphere based on the predefined rendering resolution and cone angle $\theta$. We consider the distance $\delta_n$ of the current sampling point to calculate the position $s_n$ of the next sampling point, leading to larger sampling intervals for points that are farther away, as shown in the equation:
\begin{equation}
    s_n=\sum_{k=0}^{n}{\delta_{k}}, \quad
    \delta_n=
    \begin{cases}
        c_0,  & n=0 \\
        c\cdot s_{n-1}\tan{\theta},  & n \ge1
    \end{cases}
    \label{eq: sampling dist}
\end{equation}
where $c$ controls the rate of increase in the sampling step size. We use octree nodes of depth $d_n=\lceil{\log_2{\delta_n}/{l_{0}}}\rceil$ in the octree structure for sampling, where $l_0$ denotes the minimum side length of the leaf node in the octree. This significantly reduces the number of necessary samples and accelerates computations, while still maintaining the final rendering quality. Based on these processes, we can formulate the new rendering equation as follows:
\begin{equation}
\begin{aligned}
    L(s)=\sum_{n=1}^N w_n C_n^{d_n} T_n\left(1-e^{-\sigma_n^{d_n} \delta_n}\right) 
\end{aligned}
\label{eq: new rendering disc}
\end{equation}
where $C_n^{d_n}$ and $\sigma_n^{d_n}$ represents the radiance and density of sampling point $n$ at a depth of $d_n$. And $w_n=\delta_n/s_n$ is the weight for each sampling point, while $T_n=\exp{(-\sum_{k=1}^{n-1} \sigma_k^{d_k} \delta_k)}$ is the transmittance which represents the degree of light attenuation in medium. To ensure the differentiability of the above process, we derive the derivatives of the color and opacity with respect to the forward rendering process. Please see details in supplementary file.

\begin{figure}[htb]
    \centering
    \includegraphics[width=\linewidth]{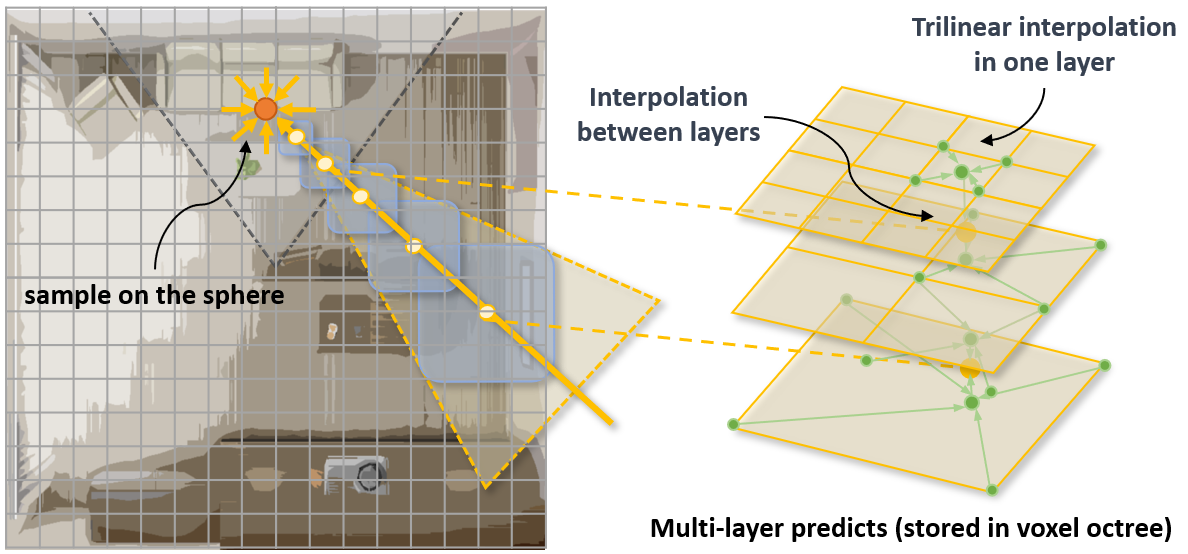}
    \caption{Rendering layer.}
    \label{fig:rendering_layer}
\end{figure}

\subsection{Loss Function}\label{sec:Loss Function}
Our model is trained using synthetic data with ground truth $\{I, D, O_s, \{I_\text{nv}, D_\text{nv}, P_\text{nv}\}_N\}$, where $I$ is the input image, $D$ denotes pixel-wise depth, $O_s$ is a voxel octree constructed from a complete scene point cloud, and $\{I_\text{nv}, D_\text{nv}, P_\text{nv}\}_N$ represent HDR panoramas, depth panoramas, and camera poses for $N$ novel views, respectively. This 3D voxel octree lighting representation allows us to avoid the need for densely rendered spherical lobe lighting ground truth used in prior works \cite{ZhengqinLi.2020,ZhengqinLi.2022,RuiZhu.2022}. Additionally, this representation ensures that the estimated results maintain angular frequency and sphere coherence, similar to other volumetric representation methods \cite{ZianWang.2021, PratulP.Srinivasan.2020,JingsenZhu2022LearningbasedIR}, but with less computational overhead.

In the Depth Network, we utilize a log-encoded L2 loss to accommodate the high dynamic range of depth\cite{ZhengqinLi.2020} and a scale-invariant L2 loss to promote relative consistency, addressing the inherent scale ambiguity of depth\cite{ZianWang.2021}. 
Inspired by \cite{PratulP.Srinivasan.2020, ZianWang.2021}, we utilized an adversarial loss\cite{DBLP:conf/cvpr/IsolaZZE17} with a GAN-like structure to supervise the global lighting decoder, enhancing its ability to capture global features of the scene.

In the Light Network, to regress the ground truth volumetric fields, we utilize an octree constructed from the ground truth point cloud of the entire scene as supervision. We employ binary cross-entropy loss to determine the emptiness of a node, providing the supervision for predicting split labels:
\begin{equation}
\mathcal{L}_{\text{octree}} = \sum_d \frac{1}{N_d} \sum_{o \in O_d} \operatorname{CrossEntropy}\left(o, o_{\text{s}}\right)
\end{equation}
where $d$ represents the depth of an octree layer, $N_d$ represents the total number of nodes in the $d$-th layer, $O_d$ represents the predicted octree node status, and $o_{\text{gt}}$ is the corresponding ground truth node status.

We supervise the prediction of the illumination $\hat{L}$ using the image intensity loss of the HDR panoramic image $I_{\text{nv}}$, as the pixel values in images reflect the corresponding HDR radiance along the camera rays. To achieve this, for each pixel $p$, we calculate the camera ray starting from the camera center $\boldsymbol{c}$ along the direction $\boldsymbol{r}$ using the camera pose $P_{\text{nv}}$ and camera intrinsic parameters. Subsequently, we render the environment map $\hat{I}_{\text{nv}, p}$ of the new view using the predicted illumination $\{\hat{L}_d\}$ with the assistance of the rendering layer:
\begin{equation}
\hat{I}_{\text{nv}, p}=\operatorname{RenderingLayer}(\boldsymbol{c}, \boldsymbol{r}, \{\hat{L}_d\})
\end{equation}
If only the alpha channel is rendered, a depth panorama $\hat{D}_{\text{nv}}$ for each camera pose can also be obtained using a similar method. We use log-encoded L2 loss to enforce consistency between the rendered new view $\hat{I}_{\text{nv}}$ and the ground truth $I_{\text{nv}}$:
\begin{equation}
\begin{aligned}
\mathcal{L}_{\text{light}} 
= & \lambda_{\text{li}}\left\|\log(I_{\text{nv}}+1)-\log(\hat{I}_{\text{nv}}+1)\right\|_2^2 \\
& + \lambda_{\text{ld}}\left\|\log(D_{\text{nv}}+1)-\log(\hat{D}_{\text{nv}}+1)\right\|_2^2
\end{aligned}
\end{equation}

An additional means of supervising the illumination is by ensuring consistency with the visible scene. So we employ the same log-encoded L2 loss $\mathcal{L}_{\text{visible}}$, which quantifies the disparity between the rendered results $\hat{I}_{\text{visible}}, \hat{I}_{\text{depth}}$ from predicted lighting voxel octree and ground truth.

\subsection{Training Details}\label{sec:Training Details}
Our model is designed for end-to-end training, employing a progressive training scheme. Initially, a single limited FOV image is used as input for the Depth Network to predict scene depth, supervised by paired depth and environment map ground truth. Subsequently, a 3D voxel octree is constructed as a new input, enabling the Light Network to predict a completed lighting voxel octree, supervised by paired HDR environment maps from novel views. Detailed information on dataset and training will follow in subsequent sections

\vspace{-0.25cm}
\paragraph{Dataset Construction}
We train our model using photorealistic renderings of indoor scenes from the FutureHouse synthetic dataset\cite{LiWHPY22}. This dataset contains artist-designed indoor panoramas with high-quality geometry and HDR environment maps, from which we extract photographs to obtain input/output pairs for training. Notably, we opted for panorama datasets over the InteriorNet dataset utilized by \cite{PratulP.Srinivasan.2020} and \cite{ZianWang.2021}, as well as the OpenRoom dataset used by \cite{ZhengqinLi.2020} and \cite{ZhengqinLi.2021}. This is because the low dynamic range of images in InteriorNet and the insufficient resolution of the panoramic data in OpenRoom, which render these datasets less suitable for lighting estimation tasks. Although the InteriorVerse proposed by \cite{JingsenZhu2022LearningbasedIR} is a better option, the environment map and other information of this dataset were unavailable at the time of writing. Fortunately, the HDR panoramas and related geometry information provided by FutureHouse allow us to construct the training data that aligns with our requirements (see the supplementary PDF for additional details).
Based on the 28,579 panoramic views from 1,752 house-scale scenes provided by FutureHouse, we construct and select 113,232 pairs of data. We use $90\%$ (1,570) of the scenes to train our model and reserve $10\%$ (180) for evaluation.

\vspace{-0.25cm}
\paragraph{Additional Details}
Our network is implemented in PyTorch \cite{NEURIPS2019_9015}, while the differentiable rendering layer is implemented using Taichi\cite{hu2019taichi}. Model training is conducted on a NVIDIA GeForce RTX 3090 GPU using the Adam optimizer\cite{KingBa15} with a batch size of 1. Given that the input image corresponds to a small proportion of the overall scene, we employ a step-by-step training scheme to ensure stable training. Initially, the Depth Network is trained separately to provide reasonable output values for the subsequent training of the Light Network, which is then trained based on real depth data and projected global features. Subsequently, both components of the network are jointly trained to address the discrepancy in the input data. This approach also bolsters the robustness of the primary Light Network across a wide spectrum of application scenarios, irrespective of whether depth is inferred or directly captured by sensors.

\section{Experiments}\label{sec:experiment}

We conduct a comprehensive evaluation of our unified lighting estimation framework by comparing it with existing methods, both qualitatively and quantitatively. Through this evaluation, we demonstrate the effectiveness and efficiency of our approach. Additionally, we evaluate our method against prior techniques in terms of lighting estimation and showcase its application in virtual object insertion. The results emphasize the effectiveness of our method in generating high-quality insertion outcomes at a lower cost.

\begin{figure*}[h]
\centering
\begin{subfigure}{0.1625\textwidth}
\captionsetup{font=tiny}
\begin{minipage}{1\linewidth}
\centering
\includegraphics[width=\linewidth]{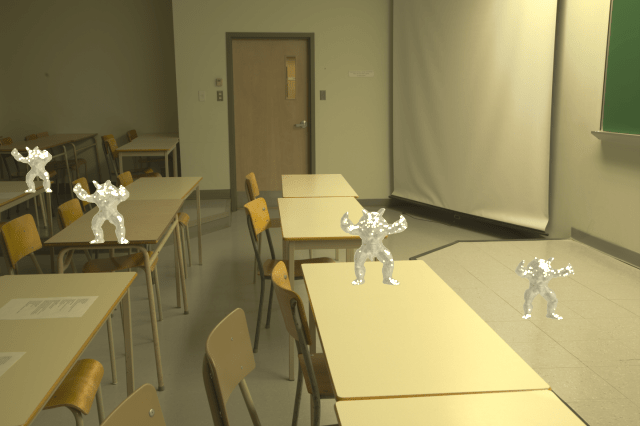}\vspace{0.1pt}
\includegraphics[width=\linewidth]{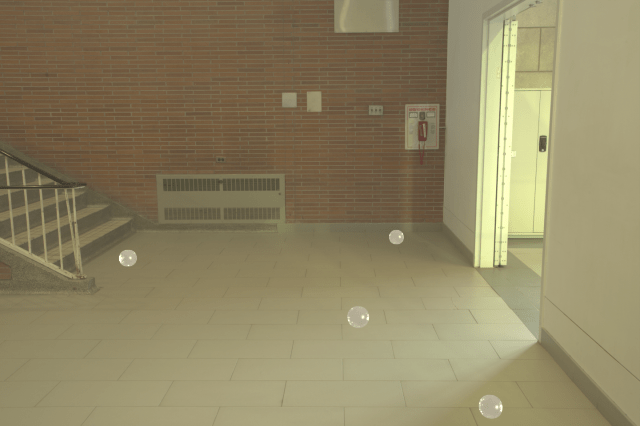}\vspace{0.1pt}
\includegraphics[width=\linewidth]{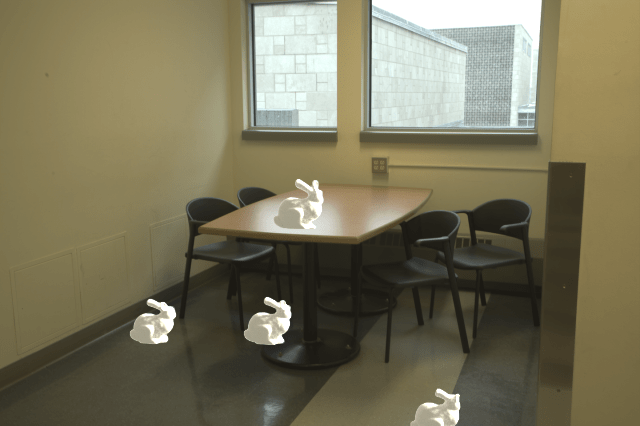}\vspace{0.1pt}
\includegraphics[width=\linewidth]{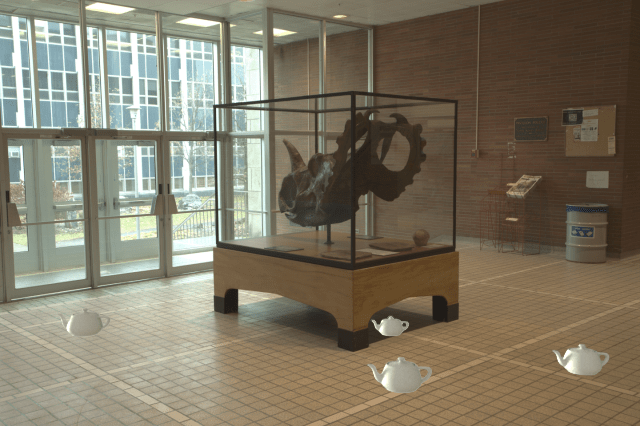}\vspace{0.1pt}
\caption{Gardner [21]}
\end{minipage}
\end{subfigure}\hspace{-1pt}
\begin{subfigure}{0.1625\textwidth}
\captionsetup{font=tiny}
\begin{minipage}{1\linewidth}
\centering
\includegraphics[width=\linewidth]{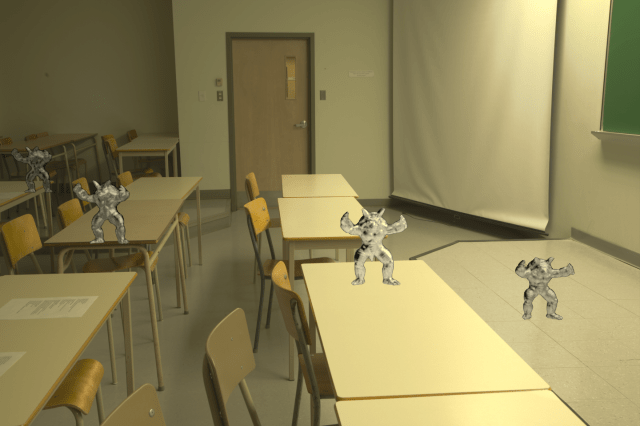}\vspace{0.1pt}
\includegraphics[width=\linewidth]{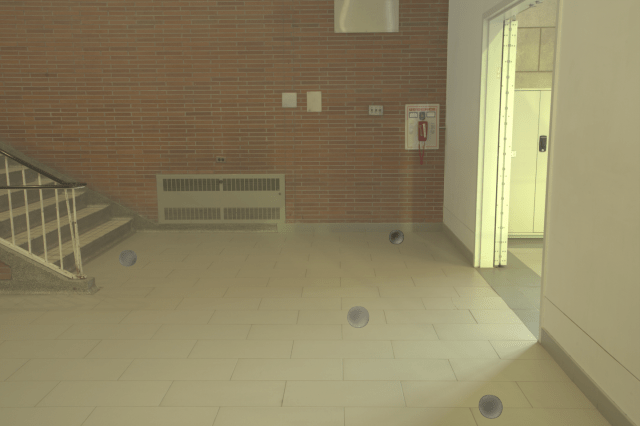}\vspace{0.1pt}
\includegraphics[width=\linewidth]{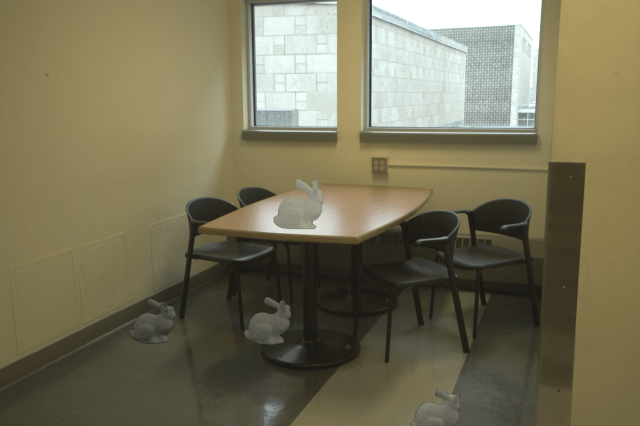}\vspace{0.1pt}
\includegraphics[width=\linewidth]{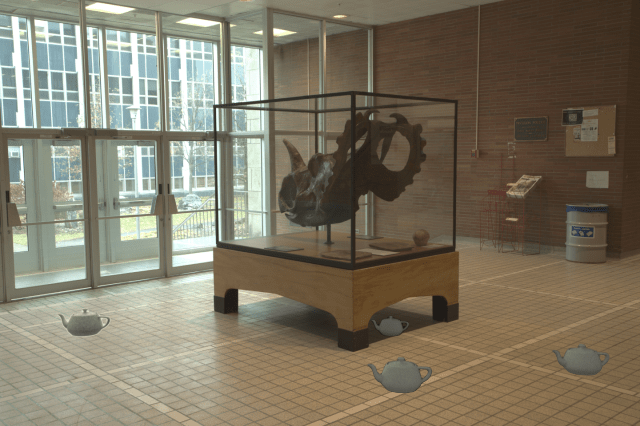}\vspace{0.1pt}
\caption{Garon [22]}
\end{minipage}
\end{subfigure}\hspace{-1pt}
\begin{subfigure}{0.1625\textwidth}
\captionsetup{font=tiny}
\begin{minipage}{1\linewidth}
\centering
\includegraphics[width=\linewidth]{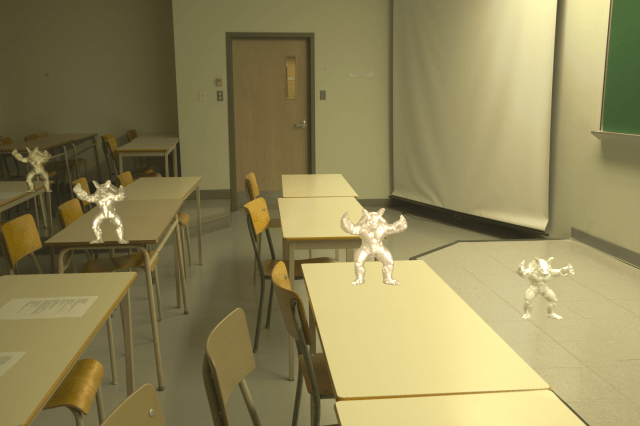}\vspace{0.1pt}
\includegraphics[width=\linewidth]{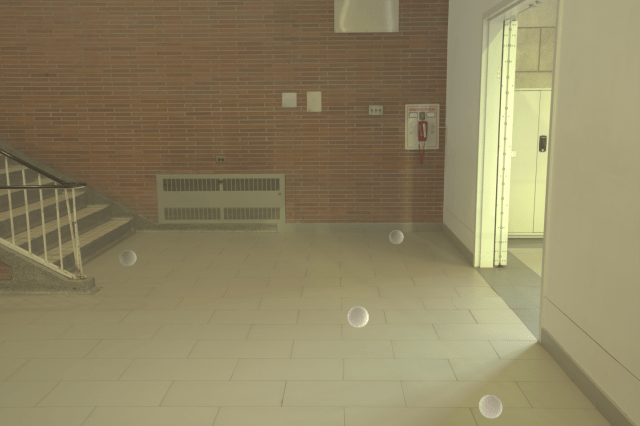}\vspace{0.1pt}
\includegraphics[width=\linewidth]{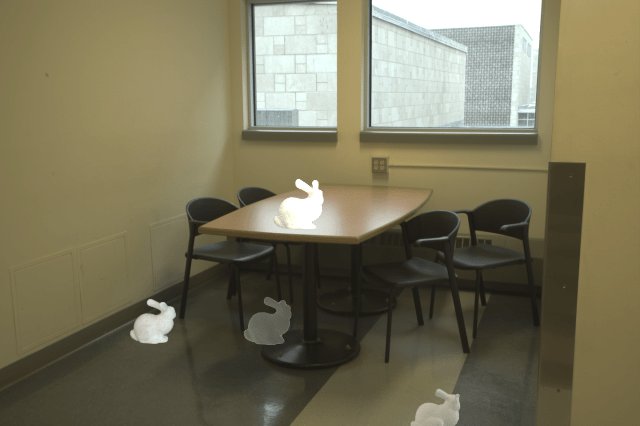}\vspace{0.1pt}
\includegraphics[width=\linewidth]{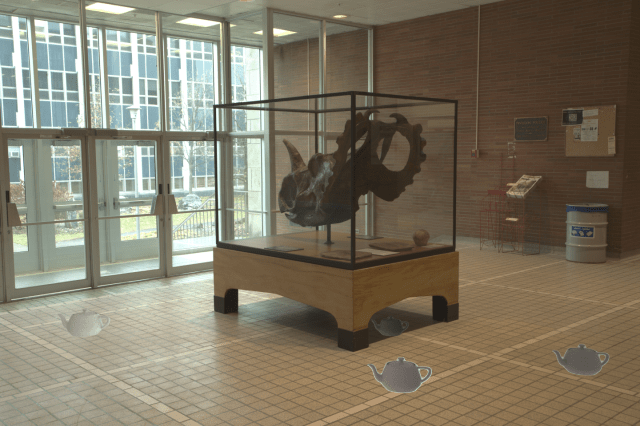}\vspace{0.1pt}
\caption{Li et al. [37]}
\end{minipage}
\end{subfigure}\hspace{-1pt}
\begin{subfigure}{0.1625\textwidth}
\captionsetup{font=tiny}
\begin{minipage}{1\linewidth}
\centering
\includegraphics[width=\linewidth]{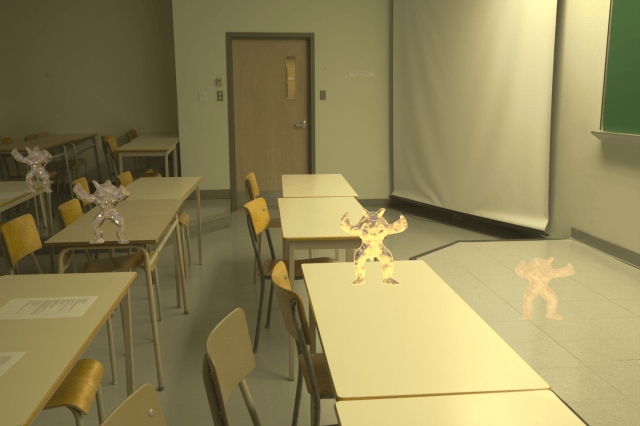}\vspace{0.1pt}
\includegraphics[width=\linewidth]{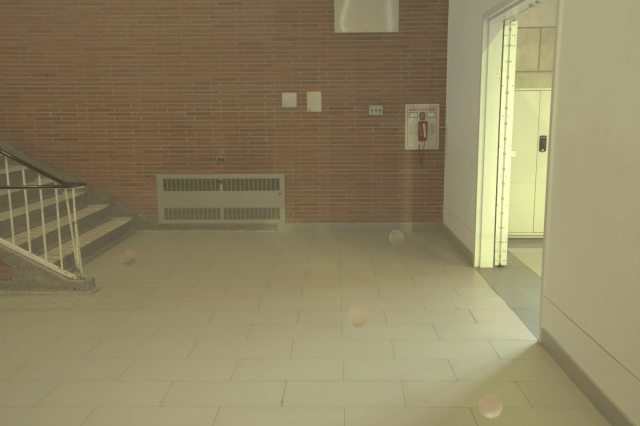}\vspace{0.1pt}
\includegraphics[width=\linewidth]{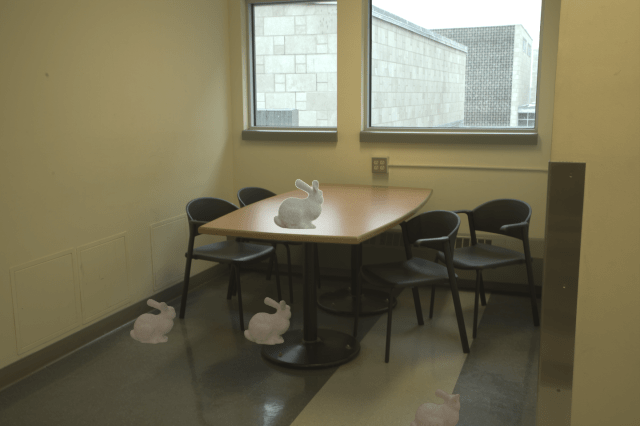}\vspace{0.1pt}
\includegraphics[width=\linewidth]{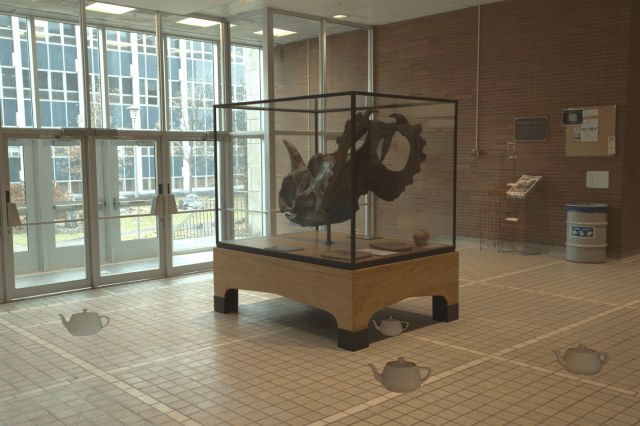}\vspace{0.1pt}
\caption{Wang [42]}
\end{minipage}
\end{subfigure}\hspace{-1pt}
\begin{subfigure}{0.1625\textwidth}
\captionsetup{font=tiny}
\begin{minipage}{1\linewidth}
\centering
\includegraphics[width=\linewidth]{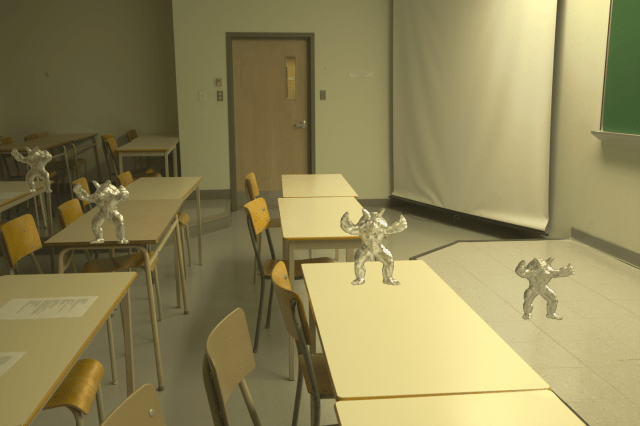}\vspace{0.1pt}
\includegraphics[width=\linewidth]{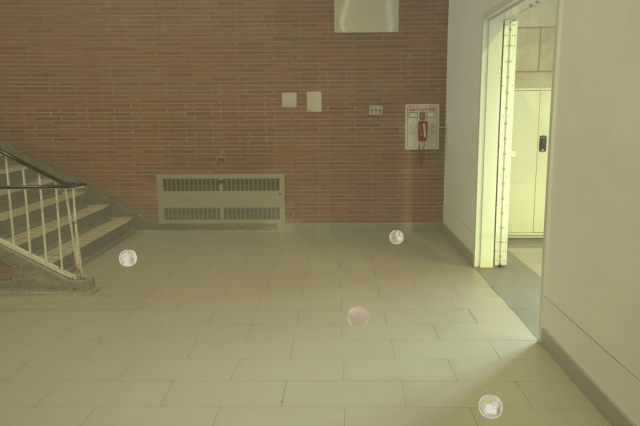}\vspace{0.1pt}
\includegraphics[width=\linewidth]{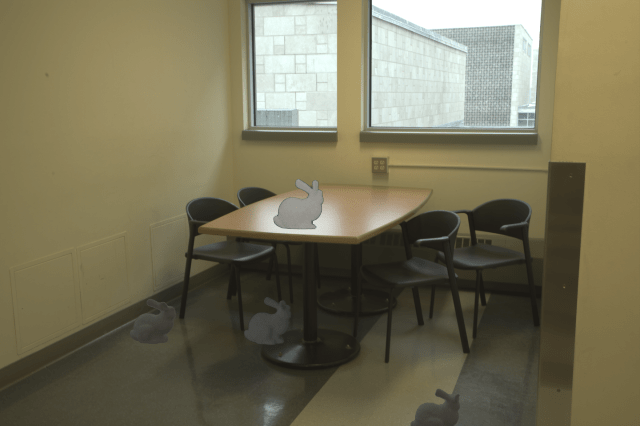}\vspace{0.1pt}
\includegraphics[width=\linewidth]{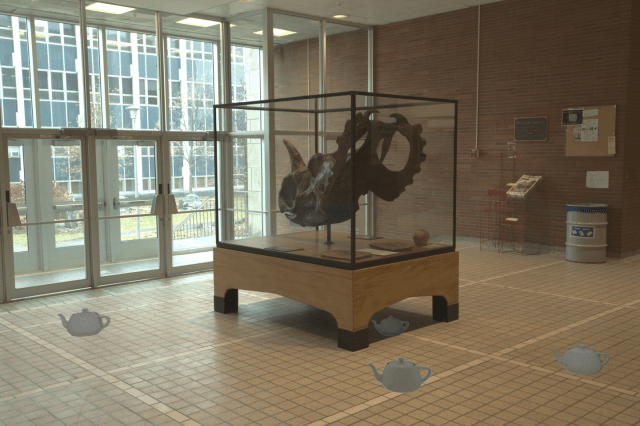}\vspace{0.1pt}
\caption{Ours}
\end{minipage}
\end{subfigure}\hspace{-1pt}
\begin{subfigure}{0.1625\textwidth}
\captionsetup{font=tiny}
\begin{minipage}{1\linewidth}
\centering
\includegraphics[width=\linewidth]{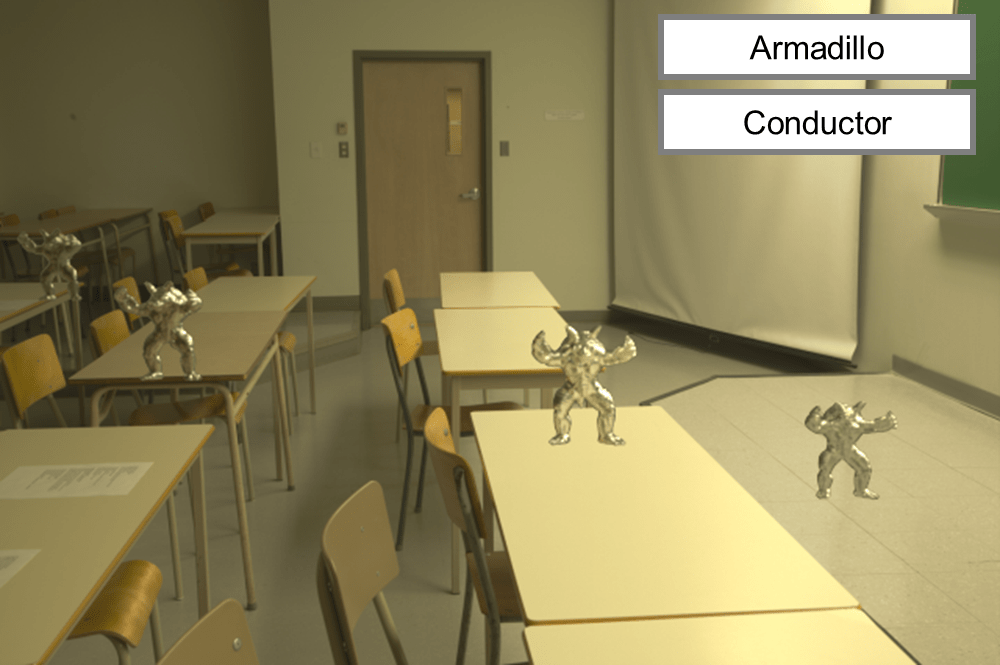}\vspace{0.1pt}
\includegraphics[width=\linewidth]{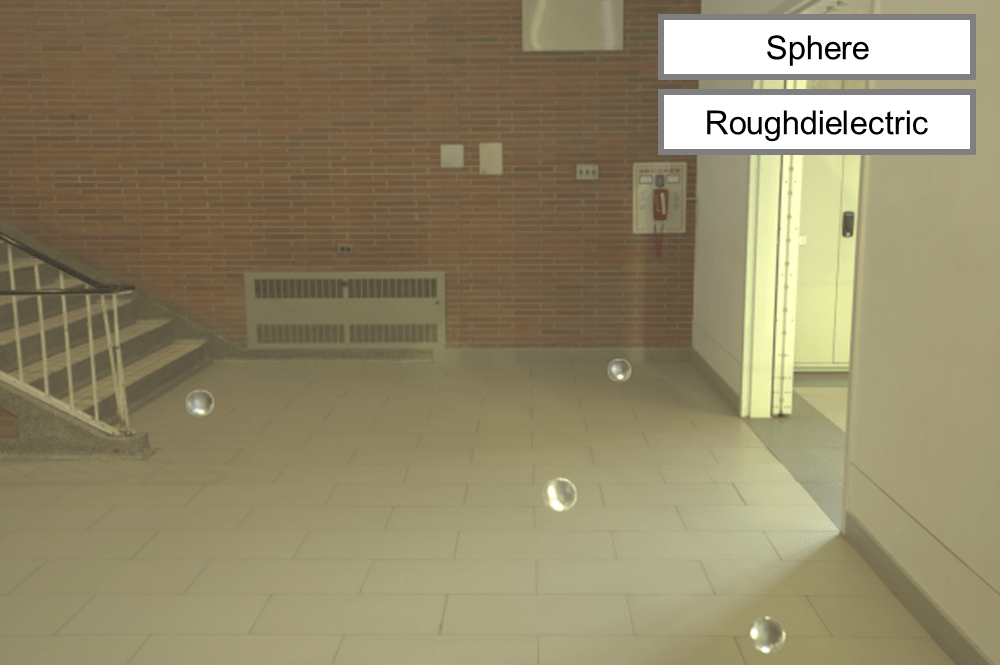}\vspace{0.1pt}
\includegraphics[width=\linewidth]{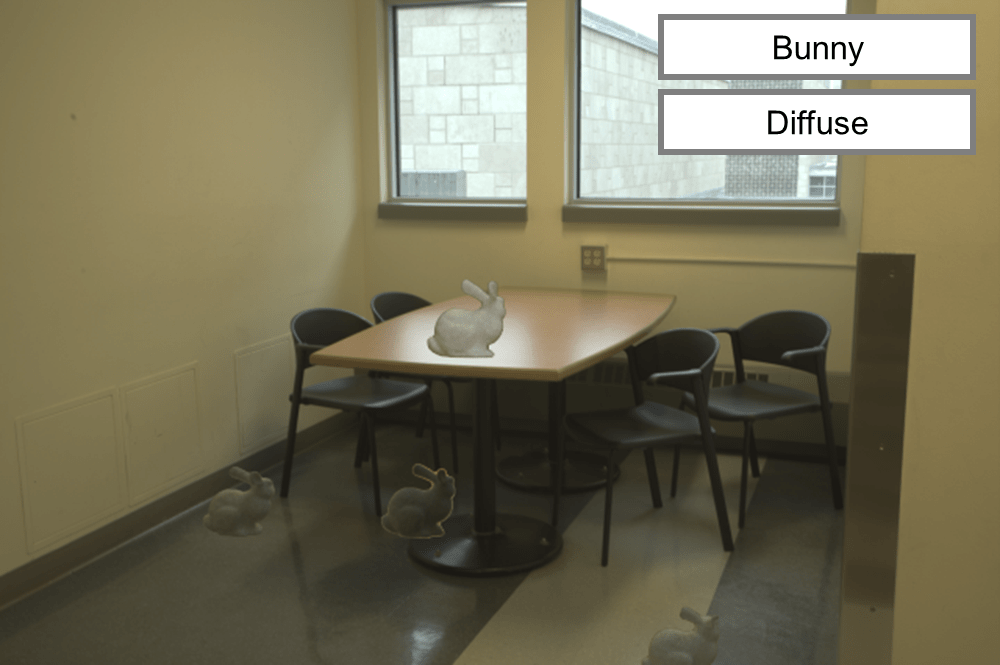}\vspace{0.1pt}
\includegraphics[width=\linewidth]{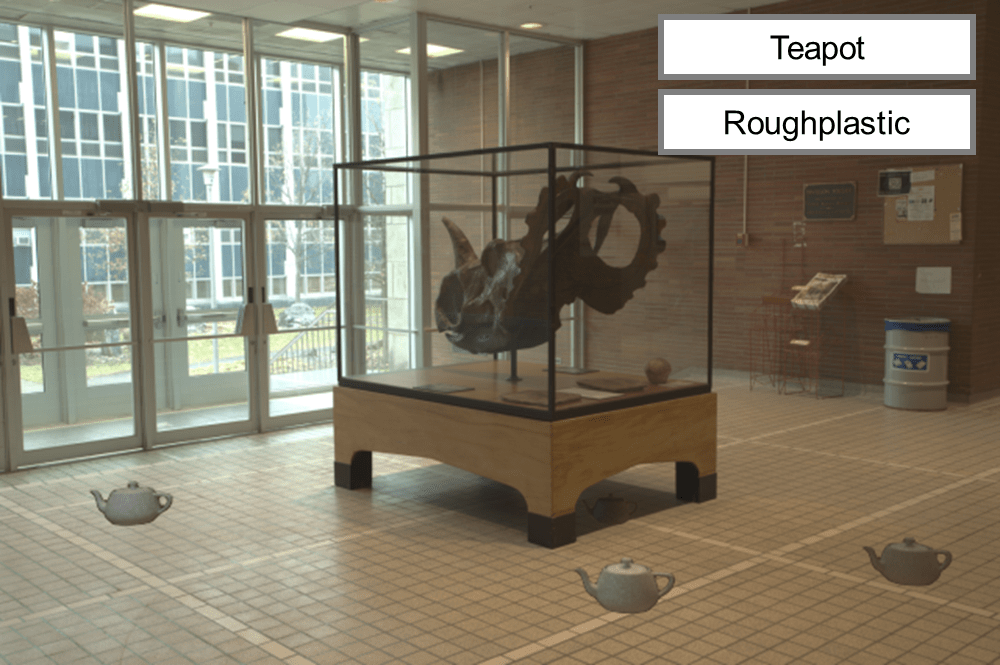}\vspace{0.1pt}
\caption{Reference}
\end{minipage}
\end{subfigure}
\caption{Qualitative evaluation on virtual object insertion.}
\label{fig:evaluation-obj-insert}
\end{figure*}

\subsection{Experiment Settings}
\paragraph{Datasets.} 
We train our network on the data constructed from the FutureHouse dataset, which was introduced in Sec.\ref{sec:Training Details}. For efficiency validation experiments, we measured inference time and the VRAM usage on a TITAN V for 480x320 input image and 256x128 output environment map resolutions. For quantitative experiments, we compared the accuracy of lighting prediction using the PSNR metric on the InteriorNet dataset. Additionally, inspired by \cite{LiWHPY22}, we compared a new metric for measuring the coherence of neighboring illumination on pairs built from the FutureHouse dataset.
For qualitative comparison experiments (perhaps the most effective lighting evaluation), we compared with prior works by visualizing object insertion results at a given location on the Laval Indoor dataset\cite{MarcAndreGardner.2017}. Following \cite{MarcAndreGardner.2017, MathieuGaron.2019, Zhao.2021}, we also designed a user study to validate that our spatially-coherent lighting estimation and object rendering method can achieve more photorealistic effects compared to other methods. 

\vspace{-0.25cm}
\paragraph{Baselines.}
In our evaluation, we compared our method with the current state-of-the-art techniques in single RGB indoor photograph-based light estimation, considering both quantitative and qualitative aspects. For the efficiency comparison experiment, we compared \cite{ZhengqinLi.2020} and several algorithms that also utilize three-dimensional neural networks to predict three-dimensional lighting representations, including \cite{PratulP.Srinivasan.2020, ZianWang.2021, JingsenZhu2022LearningbasedIR}. For the quantitative experiment, we compared the light estimation models including \cite{MarcAndreGardner.2017, MathieuGaron.2019, SoumyadipSengupta.2019, ChloeLeGendre.2019, ZhengqinLi.2020, PratulP.Srinivasan.2020,  ZianWang.2021}. For the qualitative experiment and user study experiment, we compared \cite{MarcAndreGardner.2017, SoumyadipSengupta.2019, ZhengqinLi.2020, ZianWang.2021, PratulP.Srinivasan.2020}.

\vspace{-0.25cm}
\paragraph{Virtual Object Insertion Setup.}
To ensure a fair comparison across as many methods as possible, we uniformly adopt Image-Based Lighting (IBL) for rendering. This involves illuminating the virtual objects using environment maps output by different methods (or environment maps converted from the output). Furthermore, for our method, we have implemented an efficient, high-quality fusion rendering approach that combines point clouds, mesh, and lighting voxel octree storage structures. The background scene is organized using a point cloud for simple occlusion detection; virtual objects are represented by a mesh grid, with triangular faces managed using a multi-level regular grid via SNode\cite{hu2019taichi} to accelerate the object intersection process; for each shading point on the object, the radiance is computed using the cone tracing method introduced earlier, sampling from our predicted lighting voxel octree. For further details, please refer to the supplementary file.

\subsection{Comparisons to Baseline Methods}
\paragraph{Quantitative evaluation on InteriorNet dataset.} 
We compared the lighting prediction performance of our method with baseline methods on the InteriorNet synthetic indoor dataset, and the results are shown in the Table.\ref{tab:evaluation InteriorNet}. Our method outperforms both Gardner et al. \cite{MarcAndreGardner.2017} and NIR \cite{SoumyadipSengupta.2019} significantly, as it can better model spatially-varying lighting. Compared to spatially-varying method such as DeepLight \cite{ChloeLeGendre.2019}, Garon et al. \cite{MathieuGaron.2019} and Li et al.\cite{PratulP.Srinivasan.2020}, our method still be better. Our method is slightly inferior to Lighthouse\cite{PratulP.Srinivasan.2020}, Wang et al.\cite{ZianWang.2021} in terms of accuracy. It is worth noting that Lighthouse\cite{PratulP.Srinivasan.2020} uses stereo pairs as input, which provides more information about depth and visible surfaces compared to monocular images. On the other hand, Wang et al.\cite{ZianWang.2021} 
is based on complex inverse rendering frameworks, which provide more prior knowledge about materials and geometry.
Fig.\ref{fig:experiment_lightweight} illustrates our comparative experiment on efficiency, demonstrating that our method estimates spatially-coherent lighting at interactive frame rates with minimal cost.

\begin{table}[htb]
    \centering
    \resizebox{1.0\columnwidth}{!}{
    \begin{tabular}{
        m{0.25\textwidth}<{\centering}|m{0.25\textwidth}<{\centering}}
    \toprule
        Method & PSNR(dB)$\uparrow$ \\
    \midrule
        Gardner et al. \cite{MarcAndreGardner.2017} & 13.42 \\
        NIR\cite{SoumyadipSengupta.2019} & 16.59 \\
        DeepLight \cite{ChloeLeGendre.2019} & 13.36 \\
        Garon et al. \cite{MathieuGaron.2019} & 13.21 \\
        Li et al.\cite{ZhengqinLi.2020} & 16.66 \\
        Lighthouse\cite{PratulP.Srinivasan.2020} & 17.29 \\
        Wang et al.\cite{ZianWang.2021} & \textbf{17.37}\\
        Ours & 16.82 \\
    \bottomrule
    \end{tabular}}
    \captionof{table}{Quantitative evaluation on InteriorNet dataset.}
    \label{tab:evaluation InteriorNet}
\end{table}

\vspace{-0.25cm}
\paragraph{Quantitative evaluation on FutureHouse dataset.}
We further evaluated the spatial consistency of estimated lighting variations on the FutureHouse dataset, as shown in Table \ref{tab:user-study}. We created test samples from this dataset, each comprising two input images and $N$ ground truth environment maps, which represent lighting at a randomly chosen 3D location within the camera frustum of the input image. For the same input scene, we estimated a set of environment maps $\tilde{L}$ at $N$ positions. Drawing inspiration from the SC loss proposed in PhyIR\cite{LiWHPY22}, we devised a new metric to quantitatively assess the spatially-coherent quality of estimations from different methods and computed:
\begin{equation}
    L_{SC}=\frac{1}{N} \sum\left(|\tilde{L}-L| \odot e^{\alpha\|\nabla {D}\|_1}\right),
\end{equation}
where $\nabla \tilde{D}$ is the gradient of the panorama depth. The exponential function re-weights the metrics of neighboring light probes based on the depth gradient. It can be seen that compared to using single or pixel-wise environment map estimation algorithms such as Gardner et al.\cite{MarcAndreGardner.2017} and Li et al.\cite{ZhengqinLi.2020}, our algorithm shows significant improvement in this metric, thanks to the 3D lighting voxel octree and rendering lay we designed. Compared to methods that also use 3D lighting representation like Lighthouse\cite{PratulP.Srinivasan.2020}, our algorithm achieves similar accuracy. 

\vspace{-0.25cm}
\paragraph{Qualitative evaluation on virtual object insertion.}
We compared the lighting estimation and virtual object insertion results with baselines in Fig.\ref{fig:evaluation-obj-insert} on the Laval Indoor spatially-varying HDR dataset\cite{MathieuGaron.2019}. Methods \cite{MarcAndreGardner.2017, SoumyadipSengupta.2019} use a single low-resolution environment map, which cannot handle spatially-varying effects and only recovers low frequency lighting, resulting in severe artifacts. \cite{ZhengqinLi.2020, MathieuGaron.2019} employs 2D or 3D spatially-varying spherical lobes, which can produce spatially-varying lighting, but the local lighting is still low frequency spherical lobes and cannot account for angular high-frequency details. These methods fail when inserting highly specular objects. Methods \cite{PratulP.Srinivasan.2020, ZianWang.2021} use a volumetric RGB$\alpha$ lighting representation, allowing for 3D spatially-varying lighting. Our algorithm achieves similar results with lower computational cost and even better object insertion effects in some cases. Moreover, compared to the differential rendering method based on IBL lighting, the virtual objects rendered using our proposed method have more realistic reflections of the real scene.

\begin{figure}[htbp]
    \centering
    \begin{minipage}[b]{0.30\textwidth}
        \centering
        \includegraphics[width=\linewidth]{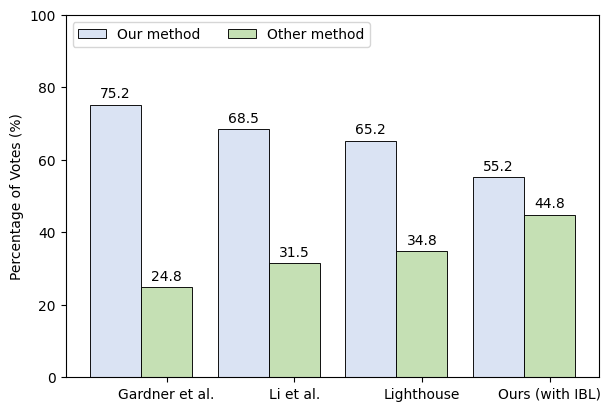}
        \caption{Pairwise user study results.}
        \label{fig:user-study-result}
    \end{minipage}
    \hfill
    \begin{minipage}[b]{0.16\textwidth}
        \centering
        \resizebox{1.0\columnwidth}{!}{
        \begin{tabular}{
            m{0.5\textwidth}<{\centering}m{0.5\textwidth}<{\centering}}
        \toprule
            Method & PSNR$\uparrow$ \\
        \midrule
            d=6 & 15.34\\
            d=6,GT & 15.56 \\
            d=6,SG & 15.62 \\
            d=7 & 17.14 \\
            d=7,GT & 17.21 \\
        \bottomrule
        \end{tabular}}
        \captionof{table}{Ablations of octree depth, GT depth map and SG on Futurehouse dataset.}
        \label{tab:evaluation Futurehouse}
    \end{minipage}
\end{figure}

\vspace{-0.25cm}
\paragraph{User study on virtual object insertion.} Similar to previous studies\cite{MarcAndreGardner.2017, MathieuGaron.2019, Zhao.2021}, we conducted a pairwise comparison user study to assess the accuracy of our light estimation method compared to ground truth. However, in order to better compare the consistency of our method’s estimation performance at different locations, we used videos instead of images as in previous studies. Participants were presented with pairs of videos containing moving virtual objects, specifically two Stanford bunnies with different materials (diffuse and metallic), composited into photo sequences. These videos were illuminated using our method and one of the state-of-the-art techniques. Participants were asked to select the image that most closely resembled the reference (ground truth), with comparisons made only between the left and right images (the middle image was provided as a reference only). Each participant provided three pairwise comparisons for each video, comparing our method against three state-of-the-art methods\cite{MarcAndreGardner.2017, ZhengqinLi.2020, PratulP.Srinivasan.2020}. A total of 20 participants (age range: 16 to 38, 5 females, 15 males) took part in the study. 
Statistical analysis, including user responses and $p$-values (represented as $p$), was conducted using a binomial test. A summary of the statistical results, along with the corresponding p-values, is presented in Tab.\ref{tab:user-study} and Fig.\ref{fig:user-study-result}. The results indicate that our method was preferred by the majority of participants over the prior techniques across all evaluated properties, with statistical significance (p$<$0.001).

\begin{table}[htbp]
    \centering
    \resizebox{1.0\columnwidth}{!}{
    \begin{tabular}{
        m{0.15\textwidth}<{\centering}m{0.11\textwidth}<{\centering}m{0.12\textwidth}<{\centering}m{0.1\textwidth}<{\centering}}
        \toprule
        Ours vs. -  & SC Metric$\downarrow$ & Response$\uparrow$ & p-value$\downarrow$ \\
        \midrule
        Gardner et al.\cite{MarcAndreGardner.2017}& 0.291  & $75.2\%$vs.$24.8\%$ & $<$0.001 \\
        Li et al.\cite{ZhengqinLi.2020} & 0.274 & $68.5\%$vs.$31.5\%$  & $<$0.001 \\
        Lighthouse \cite{PratulP.Srinivasan.2020}& 0.243  & $65.2\%$vs.$34.8\%$ & $<$0.001 \\
        Ours & \textbf{0.226} & -  & - \\
        \bottomrule
    \end{tabular}}
    \caption{Statistical results of quantitative evaluation for spatically-coherent metrics and user study on FutureHouse dataset.}
    \label{tab:user-study}
\end{table}

\vspace{-0.25cm}
\paragraph{Ablation evaluation on the FutureHouse dataset.} A deeper octree or more complex models may potentially augment the PSNR of our method's predictions, but such enhancements necessitate trade-offs. We have undertaken additional ablation studies, exploring varying octree depths and more sophisticated lighting parametric models (\textit{i.e.}, Spherical Gaussian \cite{ZianWang.2021}). These studies, as delineated in Table \ref{tab:evaluation Futurehouse}, provide partial corroboration of our perspective.

\section{Conclusion}
\label{sec:conclusion}
In conclusion, our study introduces a novel framework for efficient and high-quality estimation of spatially-coherent indoor lighting from a single image. Leveraging voxel octree and a lightweight lighting estimation network with a multi-scale rendering layer, our approach significantly reduces memory and computational resources while ensuring accurate and 3D spatially-coherent lighting estimation. These advancements present promising practical applications for more realistic AR/MR experiences. Regrettably, our approach currently falls short of SOTAs in metrics such as PSNR. Future work will aim to incorporate advanced generative models to improve our method. Another potential direction is to introduce joint estimation tasks like inverse rendering to strike a better balance between performance and accuracy, fully capitalizing on the octree-based structure.

\paragraph{Acknowledgements.} 
This work was supported by the National Nature Science Foundation of China under 62272019.

{\small
\bibliographystyle{ieeenat_fullname}
\bibliography{11_references}
}

\ifarxiv \clearpage \appendix \input{12_appendix} \fi

\end{document}